\pdfoutput=1

\documentclass[11pt]{article}

\usepackage[final]{coling}

\usepackage{times}
\usepackage{latexsym}

\usepackage[T1]{fontenc}

\usepackage[utf8]{inputenc}

\usepackage{microtype}

\usepackage{inconsolata}

\usepackage{graphicx}

\usepackage{multirow}
\usepackage{CJKutf8}
\usepackage{arydshln} 
\usepackage{amssymb}

\usepackage{pifont}

\usepackage{enumitem}

\definecolor{DarkGreen}{RGB}{0,100,0}
\definecolor{DarkYellow}{rgb}{0.8, 0.8, 0.0}
\definecolor{DarkBrown}{rgb}{0.4, 0.2, 0.1}
\definecolor{DarkBlue}{rgb}{0.0, 0.0, 0.5}
\definecolor{DarkRed}{rgb}{0.5, 0.0, 0.0}
\usepackage{amsmath}
\usepackage{marvosym}

\title{SILC-EFSA: Self-aware In-context Learning Correction for Entity-level Financial Sentiment Analysis}
\author{
 \textbf{Senbin Zhu\textsuperscript{1}\footnotemark[1]},
 \textbf{Chenyuan He\textsuperscript{1}\thanks{Equal contribution}},
 \textbf{Hongde Liu\textsuperscript{1}},
 \textbf{Pengcheng Dong\textsuperscript},
 \textbf{Hanjie Zhao\textsuperscript{1}},
\\
 \textbf{Yuchen Yan\textsuperscript{1}},
 \textbf{Yuxiang Jia\textsuperscript{1}\thanks{Corresponding author}},
 \textbf{Hongying Zan\textsuperscript{1}},
 \textbf{Min Peng\textsuperscript{2}}
\\
 \textsuperscript{1}School of Computer and Artificial Intelligence, Zhengzhou University, China
\\
 \textsuperscript{2}School of Computer Science, Wuhan University, China
\\
\text{\{nlpbin,hechenyuan\_nlp,lhd\_1013,dongpc\}@gs.zzu.edu.cn, pengm@whu.edu.cn}
\\
 \small{
   \textbf{Correspondence:} \href{ieyxjia@zzu.edu.cn}{ieyxjia@zzu.edu.cn}
 }
}

\begin{document}
\maketitle
\begin{abstract}

In recent years, fine-grained sentiment analysis in finance has gained significant attention, but the scarcity of entity-level datasets remains a key challenge. To address this, we have constructed the largest English and Chinese financial entity-level sentiment analysis datasets to date. Building on this foundation, we propose a novel two-stage sentiment analysis approach called Self-aware In-context Learning Correction (SILC). The first stage involves fine-tuning a base large language model to generate pseudo-labeled data specific to our task. In the second stage, we train a correction model using a GNN-based example retriever, which is informed by the pseudo-labeled data. This two-stage strategy has allowed us to achieve state-of-the-art performance on the newly constructed datasets, advancing the field of financial sentiment analysis. In a case study, we demonstrate the enhanced practical utility of our data and methods in monitoring the cryptocurrency market. Our datasets and code are available at \url{https://github.com/NLP-Bin/SILC-EFSA}.


\end{abstract}
\section{Introduction}

The importance of sentiment analysis in the financial domain has increasingly become apparent. As early as 1970, Fama recognized the potential of sentiment analysis in finance and introduced the concept of the Efficient Market Hypothesis (EMH)~\cite{fama1970efficient}. The EMH suggests that stock prices respond to unexpected fundamental information, supporting the use of sentiment analysis in finance. With the rapid growth of the internet and the financial sector, numerous stock reports, research papers, and investor opinions have become valuable for assessing companies and events, playing a key role for both investors and regulators.


\begin{figure}[t]
  \includegraphics[width=\columnwidth]{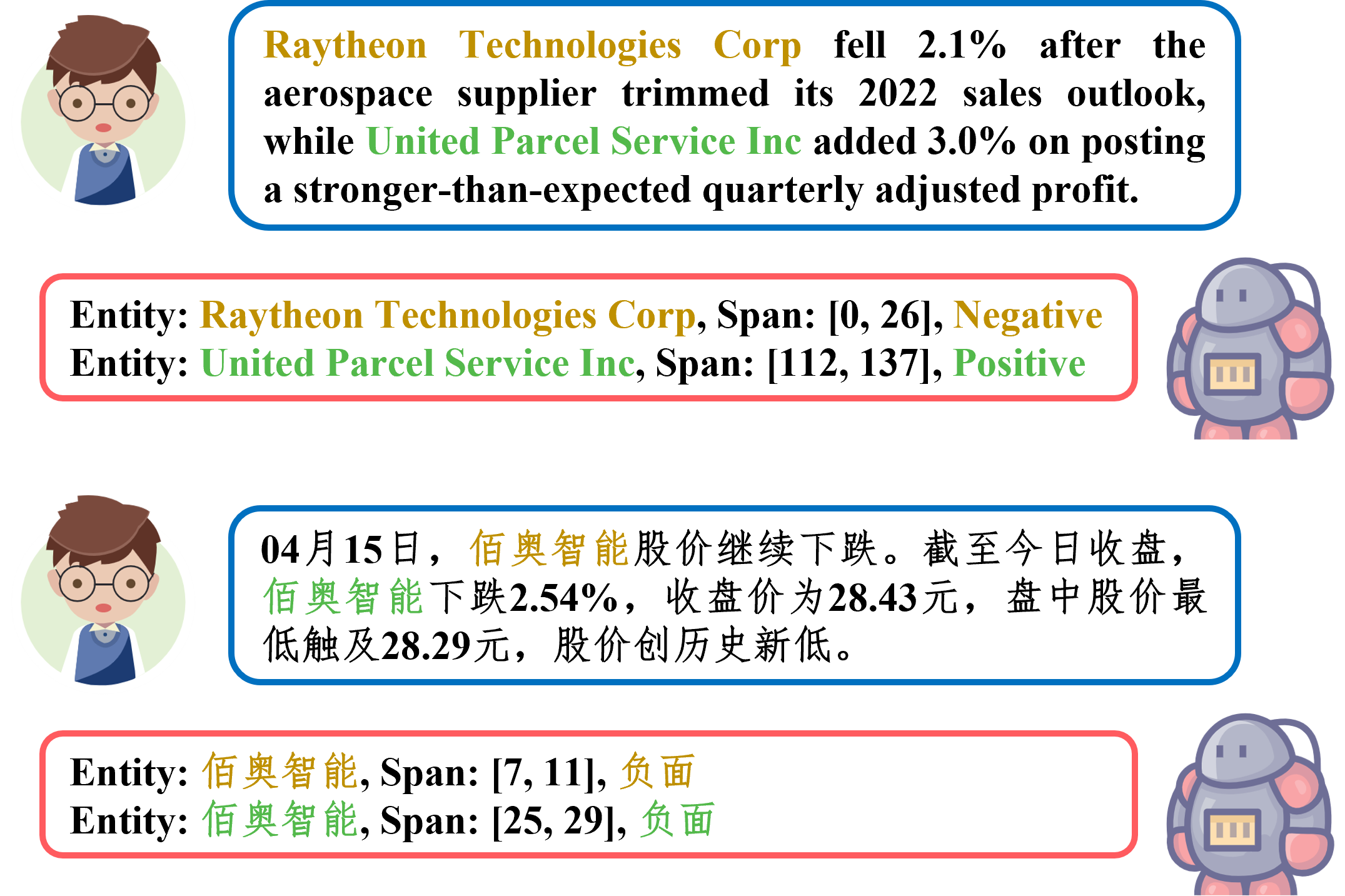}
  \caption{Examples of financial entity-level sentiment analysis data in English and Chinese.  
  In the task, the objective is to identify financial entities within the text and analyze their sentiment within the context. Specifically, this involves annotating each entity along with its position (span) in the text and determining its sentiment polarity (e.g., positive, negative, or neutral).}
  
  \label{fig:Example}
\end{figure}

Currently, most sentiment analysis corpora in the financial domain use sequence-level annotation. While sequence-level sentiment analysis has enhanced understanding of financial dynamics, many financial texts (such as news articles, analyst reports, and social media data) often contain multiple entities with differing sentiments~\cite{malo2014good,huang2023finbert,sinha2021impact,shah2023trillion}. 
Consequently, the need for fine-grained entity-level sentiment analysis in the financial domain has emerged. This task requires extracting the entities in the text and their location information, as well as the sentiment polarity corresponding to the entities.
Task examples are illustrated in Figure~\ref{fig:Example}.


~\citet{tang-etal-2023-finentity} construct the first dataset with financial entity annotations and sentiment labels, named FinEntity, which clarifies the task requirements for entity-level sentiment analysis in the financial domain. However, the FinEntity dataset is relatively small in scale. To conduct a more comprehensive study of entity-level financial sentiment analysis, we thoroughly investigate current open-source financial datasets containing entity information. We select the SEntFiN dataset~\cite{sinha2022sentfin} and a Chinese financial event-level sentiment analysis dataset~\cite{chen2024efsa}, re-screen and reconstruct them, ultimately creating the largest (English and Chinese) entity-level financial sentiment analysis database to date. The construction methods will be detailed in Section~\ref{Datasets}. 

Currently, large language models (LLMs) have achieved significant research milestones in the financial domain, such as BloombergGPT~\cite{wu2023bloomberggpt} and FinLlaMA~\cite{konstantinidis2024finllama}. Most research focuses on further pre-training base models on large-scale financial data to enhance their expertise in the financial domain. However, methods based on LLMs for entity-level sentiment analysis remain relatively few. On the dataset we constructed, we design a two-stage strategy to leverage large models and achieve performance improvements.

In the first stage, we fine-tune a base model to perform entity-level sentiment analysis, and we find that its understanding of knowledge needs to be enhanced. Even after training, models still generate some incorrect predictions on the data. This is analogous to the human learning process, where initial learning often lacks comprehensive knowledge, and self-checking and error correction can further enhance performance. Based on this observation, we design an error-correction strategy in the second stage, training a self-correcting mechanism to improve model performance.
Specifically, we first obtain and filter erroneous sample data, then train a graph neural network (GNN) to retrieve relevant examples, and subsequently fine-tune the base model to function as an error-corrector to rectify the predictions from the first stage. Experiments conducted on the dataset we constructed demonstrate that this approach achieves state-of-the-art performance. Leveraging the sentiment analysis capabilities of our model, we perform information extraction on time-series financial texts from the cryptocurrency market, and achieve more accurate price predictions using an LSTM network.


The main contributions of this paper are as follows:

\noindent \textbullet\hspace{1pt} We restructure existing English and Chinese datasets to build the largest financial entity-level sentiment analysis datasets to date, providing a rich and reliable data resource for future research.

\noindent\textbullet\hspace{1pt}  We propose a novel two-stage self-correction approach, covering Initial Response Generation and Self-correction Steps, which significantly improves the model's predictive accuracy and reliability.

\noindent\textbullet\hspace{1pt}  We achieve state-of-the-art performance on our customized datasets and obtain better predictive results in real-world cases, providing substantial support for financial sentiment analysis.







\section{Related Work}

\begin{table*}[h]
  \centering
  \setlength{\tabcolsep}{3pt}
  \begin{tabular}{l c||c|c}
    \hline
    \textbf{} & \textbf{FinEntity} & \textbf{SEntFiN-Span} & \textbf{FinEntCN}\\
    \hline
    Number of Texts     &979      &10753    &10832      \\
    Single Entity Texts           &390 (39.84\%)  &7897 (73.44\%)  &8194 (75.65\%) \\
    Multiple Entity Texts         &589 (60.16\%)  &2856 (26.56\%)  &2638 (24.35\%) \\
    Average Text Length by Tokens &37.01    &9.91     &145.23      \\ 
    Max Text Length by Tokens     &300      &23       &518      \\
    Min Text Length by Tokens     &21       &3        &12      \\
    Positive Entities               &503 (23.60\%)   &5084 (35.21\%)  &8037 (53.89\%) \\
    Negative Entities              &498 (23.37\%)   &3828 (26.51\%)  &5040 (33.79\%)  \\
    Neutral Entities               &1130 (53.03\%)  &5527 (38.28\%)  &1838 (12.32\%) \\
    All Entities            &2131     &14439    &14915       \\
    Average Entity 
    Num Per Text     &2.18      &1.34    &1.38       \\
    \hline
  \end{tabular}
  \caption{\label{table data distribution}
    The statistics of the constructed datasets.
  }
\end{table*}

\subsection{Entity-level Sentiment Analysis of Financial Texts}
NLP techniques have gained widespread adoption in financial sentiment classification~\cite{kazemian2016evaluating,yang2022analyzing,chuang2022buy,xing2020financial}. Recent studies have achieved state-of-the-art performance in SemEval 2017 Task 5 and FiQA Task 1~\cite{du2023incorporating}.
With advances in NLP, sentiment analysis has shifted from coarse- to fine-grained approaches~\cite{du2024financial}. However, entity-level sentiment analysis in financial texts, a key fine-grained task, remains in its early stages~\cite{zhu-etal-2020-jie} and faces several challenges~\cite{tan2023survey}.

Existing financial sentiment classification datasets, such as Financial Phrase Bank~\cite{malo2014good}, SemEval2017~\cite{cortis2017semeval}, AnalystTone dataset~\cite{huang2023finbert}, Headline News dataset~\cite{sinha2021impact}, and Trillion Dollar Words~\cite{shah2023trillion}, are based on entire text sequences (sentences or articles). FiQA\footnote{\url{https://sites.google.com/view/fiqa/home}} is an open challenge dataset with aspect-level sentiment. However, it does not include entity annotations. For financial entity annotation datasets, FiNER~\cite{shah2023finer} and FNXL~\cite{sharma2023financial} have been created for financial entity recognition and numerical span annotation, respectively, but both lack sentiment annotations. The FinEntity dataset is a dataset with entity spans and sentiment information. 


\citet{gururangan2020don,zhang2023instruct} suggest that re-training general PLMs on domain-specific corpora enhances performance on specialized tasks. However, entity-level financial sentiment analysis requires further research due to the unique complexity of financial entities compared to general text entities~\cite{ZhangEnhancingRetrieval}.
\citet{tang-etal-2023-finentity} have achieved preliminary results in entity-level sentiment analysis tasks using a combination of FinBERT and CRF.


\subsection{LLMs in Finance}

Large Language Models (LLMs) are considered a technological breakthrough in the field of natural language processing, as exemplified by GPT-3 and GPT-4~\cite{GPT2020}. 
LLMs have been applied to various tasks in the financial domain~\cite{dredze2016twitter,araci2019finbert,bao2022plato,delucia2022bernice,konstantinidis2024finllama,ahmed2024leveraging}, from predictive modeling to generating insightful narratives from raw financial data. 

An early example of a financial LLM is BloombergGPT~\cite{wu2023bloomberggpt}. 
\citet{lan2024chinese} further apply large models to specific tasks in enterprise alert systems and constructed the FinChina SA dataset, achieving meaningful results. \citet{chen2024efsa} obtain desirable performance on event-level datasets using their proposed four-hop reasoning chains. However, it is worth noting that experiments have shown that the CoT (Chain-of-Thought) framework negatively impacts entity-level financial sentiment analysis tasks, likely due to the complex reasoning processes involved. 

Self-correction techniques aim to improve the accuracy of LLM outputs by enabling models to revise their initial predictions~\cite{pan2023automatically,kamoi2024can}. The fundamental problem with existing methods is that LLMs cannot reliably assess the correctness of their inferences~\cite{huang2024large}. Recent studies have shown that incorporating examples with feedback into the context can improve response quality~\cite{xu-etal-2024-improving}. These findings highlight the significant research potential for correction strategies in entity-level sentiment analysis using LLMs.

\section{Datasets Construction}
\label{Datasets}



\begin{figure*}[h]
  \includegraphics[width=\linewidth]{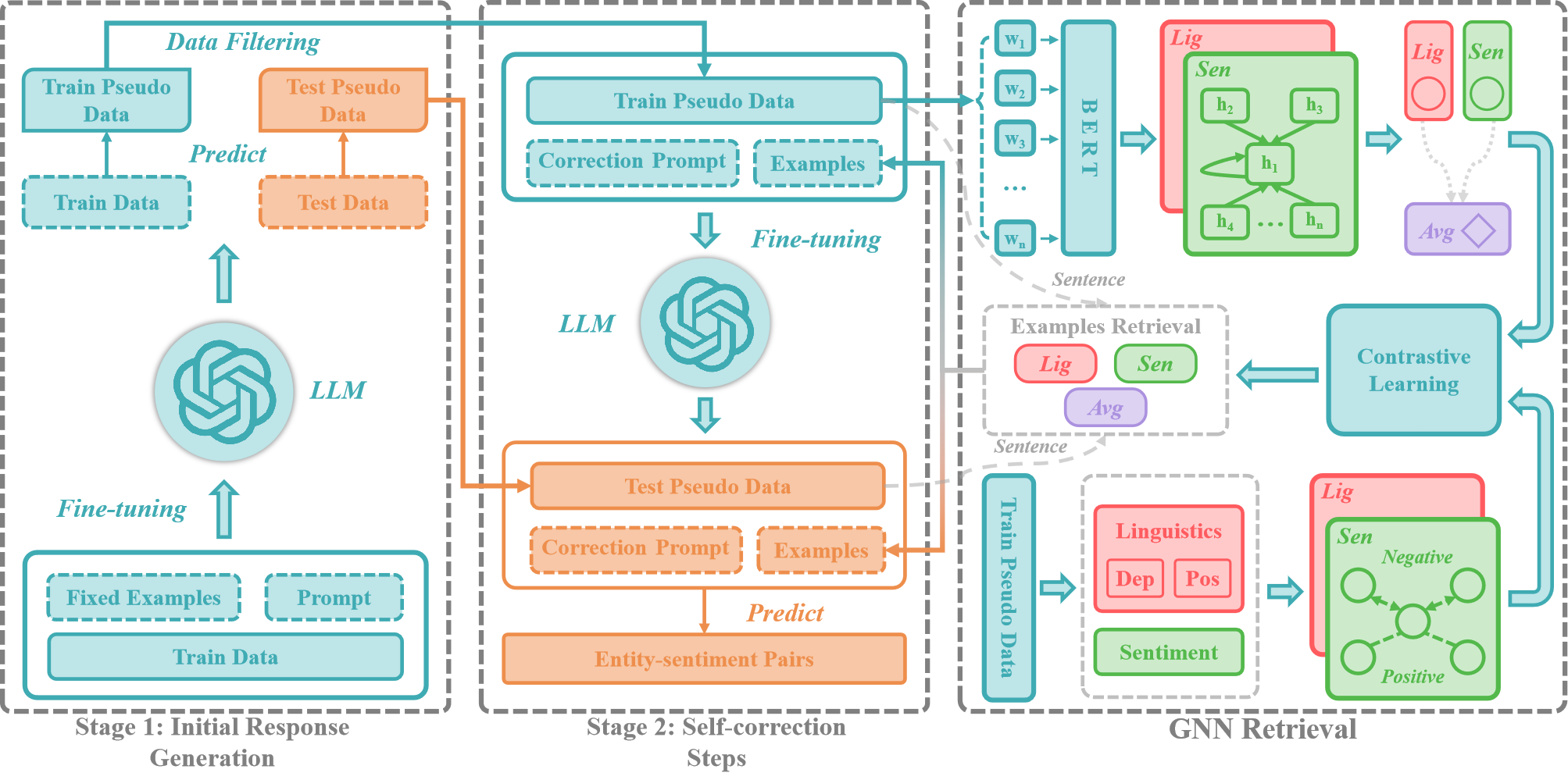} 
  \caption {Overview of our framework.}
  \label{fig:kg}
\end{figure*}

The FinEntity\footnote{\url{https://github.com/yixuantt/FinEntity}} dataset provides annotations for financial entity spans and their associated sentiments within text sequences, with sentiment labels categorized as positive, neutral, or negative~\cite{tang-etal-2023-finentity}. It treats entity-level sentiment classification as a sequence labeling task using the BILOU annotation scheme. The dataset contains 979 example sentences, with 2,131 entities in total. Additionally, approximately 60\% of the financial texts contain multiple entities. However, due to the small size of this dataset, we construct two additional datasets—one in English and one in Chinese—to provide a more comprehensive analysis of this task. Detailed information about the constructed dataset is shown in Table~\ref{table data distribution}, with 20\% of the data randomly selected as the test set for experiments.


\subsection{English Dataset}

SEntFiN\footnote{\url{https://github.com/pyRis/SEntFiN}} is a manually annotated dataset designed for fine-grained financial sentiment analysis of news headlines, with sentiment labels linked to financial entities~\cite{sinha2022sentfin}. The sentiment labels are defined as positive, neutral, and negative. 
In total, the dataset includes 14,404 entity and sentiment annotations. 

We apply a rule-based approach to add entity location tags to the annotations, aligning the label format with the FinEntity dataset for ease of subsequent work. This restructured dataset is named SEntFiN-Span.The reconstructed dataset remains relatively balanced in terms of sentiment distribution. 




\subsection{Chinese Dataset}

To address the shortage of Chinese financial sentiment analysis datasets, \citet{chen2024efsa} pioneer a new task called Event-Level Financial Sentiment Analysis, which involves predicting a five-tuple (company, department, coarse-grained event, fine-grained event, sentiment). To support this task, they construct a large-scale publicly available dataset containing 12,160 news articles and 13,725 five-tuples\footnote{\url{https://anonymous.4open.science/r/EFSA-645E}}.


To adapt this dataset for entity-level sentiment analysis tasks, we first keep data containing only single-type entity label information. Such data amount to 10,832 instances, accounting for 89\% of the dataset. We then use a rule-based approach to annotate the spans of financial entities while ignoring event labels, simplifying the task to purely entity-level financial sentiment analysis. The processed dataset has an average text length of 145.23 words, with 75.65\% of the data containing a single entity and 24.35\% containing multiple entities. 
This restructured dataset is named FinEntCN.




\section{Methodology}


The methodology of our study consists of two stages: base model fine-tuning and error correction model training. The methodological framework is illustrated in Figure~\ref{fig:kg}.

\subsection{Stage 1: Initial Response Generation}

In the first stage, we aim to fine-tune the base LLMs for entity-level sentiment analysis in the financial domain. We use two models for different datasets: LlaMA2-7b-hf-finance for the English dataset and Baichuan2-7b for the Chinese dataset. 

During the fine-tuning process, we develop multiple versions of instruction templates to ensure their general applicability. Each instruction is designed to comprehensively describe and clearly convey the task requirements, including financial entity recognition, sentiment classification, and the details of result annotation. In the output phase, the model must accurately identify entity names, clearly delineate their boundaries within the text, and classify their sentiment polarity.
To enhance the effectiveness of the fine-tuning, we incorporate three manually selected fixed examples into the instructions. These examples are carefully chosen to cover a range of scenarios within the task, providing strong representativeness and comprehensiveness. The instruction template for this stage is shown in Figure~\ref{fig:stage1prompt} in Appendix. After the fine-tuning process, we perform predictions on both the training and test sets, generating initial labels, which we refer to as pseudo-labels.

The problem can be formalized as follows:
Let $D_{train}$ and $D_{test}$ represent the training and test datasets, respectively. For each text $t\in D_{train}\cup D_{test}$, we want to predict the sentiment label $y_s$ for each entity 
$e$ within the text. The fine-tuned model $M_{ft}$ predicts the sentiment labels:
\begin{equation}
  \label{eq:stage1}
  \{(e_i, \text{start}_i, \text{end}_i, \hat{y}_i)\}_{i=1}^{n} = M_{\text{ft}}(t)
\end{equation}
Where:
 \( e_i \) represents the \( i \)-th entity recognized in the text \( t \),
 \( \text{start}_i \) and \( \text{end}_i \) denote the starting and ending positions of entity \( e_i \) in \( t \),
 \( \hat{y}_i \) is the sentiment label for entity \( e_i \) (positive, negative, or neutral),
 \( n \) is the total number of entities identified in the text.
 
The pseudo-labels are generated for both the training and test sets. Despite the higher accuracy on the training set, some pseudo-labels are erroneous, reflecting the model's limitations on this specific task. This approach provides foundational data for training the subsequent error correction model. 



\subsection{Stage 2: Self-correction Steps}

The second stage focuses on training an error correction model to identify and rectify errors in the pseudo-labels generated during the first stage. The specific steps are as follows:
We begin by filtering the pseudo-labeled data $D_{train}$. Let $C_{correct}$ and $C_{incorrect}$ represent the correctly and incorrectly labeled samples, respectively. To emphasize the model’s attention on erroneous samples, we filter the training data by retaining all incorrectly predicted samples $C_{incorrect}$, while removing a portion of the correctly predicted samples $C_{correct}$:
\begin{equation}
  \label{eq:filter}
  D_{filtered}= C_{incorrect}\cup S(C_{correct})
\end{equation}
where $S$ is a sampling function that selects a subset of correctly predicted samples.
Next, we fine-tune an error correction model $M_{correct}$ using the filtered dataset $D_{filtered}$. 

We reference the GNN-based context example retriever proposed by \citet{yang-etal-2024-faima-feature}, as shown in Figure~\ref{fig:kg}. The GNN example retriever uses a graph attention network (GAT) as the base model, with two GAT layers designed for linguistic and sentiment features. It outputs feature representations rich in linguistic and sentiment information, along with sentence-level average representations. During training, the GNN model employs contrastive learning, where linguistic features (such as syntactic dependencies and part-of-speech tags) and sentiment features (such as sentiment polarity) are extracted from the training set based on heuristic rules for comparison. This encourages the GNN's output to be optimized in both linguistic and sentiment dimensions. By encoding the dataset using the trained GNN model, three feature representations enriched with linguistic and sentiment characteristics are obtained. Finally, for prediction, approximate nearest neighbor search is used to retrieve the most similar examples from the encoded training set.

 We incorporate the retrieved examples into the context of the fine-tuning instructions, providing information on whether the pseudo-labels are correct. This allows the model to learn how to judge the accuracy of pseudo-labels. Since the model from the first stage may produce similar errors on similar examples, this retrieval method helps the correction model identify and correct these errors.
The trained error correction model is then used to detect and correct the pseudo-labels generated by the first stage on the test set. 
By employing this two-stage approach, we further enhance prediction accuracy and reliability. The corrected-labels are formally consistent with pseudo-labels.

It is important to note that during the fine-tuning of the error correction model, we do not include additional instructions related to entity-level sentiment analysis in the financial domain but focused solely on the error correction task. This design avoids potential negative impacts on model performance from the comprehensiveness of task instructions, ensuring that the model remains concentrated on the error correction task itself. The correction task prompt template can be found in Figure~\ref{fig:stage2prompt} in Appendix.

\section{Experiments}

\begin{table*}[h!]
  \centering 
  \begin{tabular}{l c|c|c|c|c|c}
    \hline
    \multirow{2}{*}{Method}    & \multicolumn{3}{c}{FinEntity} & \multicolumn{3}{c}{SEntFiN-Span} \\ \cline{2-7}
    
    &Precision	&Recall	&F1   &Precision	&Recall 	&F1\\
    \hline
    BERT                &-      &-     &0.8000*      &-     &-      &-  \\
    BERT-CRF            &-      &-     &0.8100*     &-     &-      &-  \\
    FinBERT             &-      &-     &0.8300*      &-     &-      &-  \\
    FinBERT-CRF         &-      &-     &0.8400*      &-     &-      &-  \\
    SpanABSA            &0.6635  &0.5210  &0.5837   &0.7922     &0.7029    &0.7449    \\
    T5-base             &0.8578  &0.8477  &0.8527  &0.7639  &0.7546  &0.7592  \\
    BGCA         &0.8625  &0.8524  &0.8574  &0.7626  &0.7670  &0.7648  \\
    InstructABSA        &0.8152  &0.7471  &0.7797   &0.7881 &0.7826 &0.7853  \\ 
    LlaMA2-7B           &0.8920  &0.8237 &0.8565  &0.7677   &0.7719   &0.7698  \\\hline
    GPT-3.5$_{(0-shot)}$  &0.5253  &0.6723  &0.5902   &0.5143 &0.6923 &0.5902  \\
    GPT-3.5$_{(3-shot)}$     &0.7265  &0.6347  &0.6775   &0.5644 &0.6525 &0.6053  \\
    GPT-4o$_{(0-shot)}$     &0.7989  &0.6714  &0.7296   &0.5354 &0.6721 &0.5952  \\
    GPT-4o$_{(3-shot)}$      &0.7751  &0.7588  &0.7669   &0.6679 &0.6299 &0.6484  \\ \hline
    
   Stage 1 LlaMA2-7B-Finance  &0.8983  &0.8399   &0.8681    &0.7833   &0.7789   &0.7811 \\
    \hline\hline
   Stage 2 (fix)      &0.9082    &0.8632   &0.8826   &0.7881  &0.7887  &0.7884   \\
   Stage 2 (gnn)      &\textbf{0.9104}   &\textbf{0.8724}   &\textbf{0.8910}   &\textbf{0.7895}  &\textbf{0.7898}  &\textbf{0.7896}  \\
    \hline
  \end{tabular}
  \caption{\label{table FinEntity-SEntFiN-result}
    The experimental results on two English datasets (FinEntity, SEntFiN-Span). The BGCA’s base model is T5. The reported scores are F1 scores over three runs. ‘-’ denotes that the corresponding results are not available. ‘*’ Indicates that the results are derived from previous studies. The best results are bolded. ‘fix’ indicates that the in-context examples used are fixed, while ‘gnn’ indicates they are retrieved using the GNN-based retriever.
  }
\end{table*}

\begin{table}[h!]
  \centering
  \begin{tabular}{l c|c|c}
    \hline
    \multirow{2}{2cm}{Method}    & \multicolumn{3}{c}{FinEntCN}   \\ \cline{2-4}
    &Precision	&Recall	&F1  \\
    \hline
    Ch\_finT5\_base     &0.3582 &0.3627 &0.3604      \\
    BGCA         &0.3712       &0.3804     &0.3757      \\
    \hline
    {GPT-3.5$_{(0-shot)}$} 
     &0.1564 &0.3723 &0.2203 \\
    
    {GPT-3.5$_{(3-shot)}$}   
     &0.4281 &0.4858 &0.4551 \\
     {GPT-4o$_{(0-shot)}$}   
     &0.2097 &0.4468 &0.2854 \\
    {GPT-4o$_{(3-shot)}$}  
     &0.2908 &0.5177 &0.3724 \\
    \hline
    Stage 1        &0.8574      &0.8546     &0.8560      \\
    \hline\hline
    Stage 2 (fix)           &0.8625      &0.8582     &0.8604            \\
    Stage 2 (gnn)           &\textbf{0.8671}     &\textbf{0.8681}     &\textbf{0.8675}      \\
    \hline
  \end{tabular}
  \caption{\label{table FinEntCN result}
    The experimental results on the Chinese dataset FinEntCN. The Stage 1 results are obtained by fine-tuning Baichuan2-7B-Base-LLaMAfied, and BGCA's base model is Ch\_finT5\_base.
  }
\end{table}

\subsection{Baselines}

FinBERT-CRF: 
FinBERT\footnote{\url{https://huggingface.co/yiyanghkust/finbert-tone}} is a BERT variant for finance, and FinBERT-CRF adds a CRF layer for token label dependencies~\cite{yang2020finbert, tang-etal-2023-finentity}.

SpanABSA: SpanABSA is a span-based extract-then-classify framework~\cite{hu2019open}.

Instruct ABSA: InstructABSA~\cite{scaria2024instructabsa} is an aspect-based sentiment analysis method based on instruction learning. 

T5: T5 is a general text generation model proposed by Google Research~\cite{2020t5}. 

Ch\_finT5\_base\footnote{\url{https://huggingface.co/SuSymmertry/BBT/tree/main/Model/1b}}: Pre-trained Language Model in the Chinese financial domain~\cite{lu2023bbt}.

BGCA: BGCA is a unified bidirectional generation framework for cross-domain ABSA tasks~\cite{deng2023bidirectional}. 

LlaMA2-7B-Finance\footnote{\url{https://huggingface.co/cxllin/Llama2-7b-Finance}}: This model is fine-tuned on a financial dataset based on the LlaMA2-7B language model. 

Baichuan2-7B-Base-LLaMAfied\footnote{\url{https://huggingface.co/hiyouga/Baichuan2-7B-Base-LLaMAfied}}: This is the LLaMAfied version of the Baichuan2-7B-Base model~\cite{huang2024c}. 


ChatGPT: To compare with state-of-the-art generative LLMs, we use the OpenAI API\footnote{\url{https://platform.openai.com}} to evaluate ChatGPT's zero-shot and few-shot in-context learning performance.  Detailed prompts are provided in Appendix ~\ref{Appendix A}.











\subsection{Experimental Settings}
The detailed experimental settings of our method and parameter configurations of methods such as SpanABSA, BGCA, and InstructABSA can be found in Appendix~\ref{Appendix B}.



\subsection{Overall Performance}

    

Table~\ref{table FinEntity-SEntFiN-result} and Table~\ref{table FinEntCN result} present the main experimental results, demonstrating that our proposed method outperforms all benchmark models on most metrics across three datasets. On the FinEntity dataset, our approach improves the F1 score by 5.1\% compared to the previous best method, and it also performs well on the other two datasets. In our comparative study, we explore pre-trained models that have been successfully applied to aspect-based sentiment analysis tasks. The results show that the fine-tuning approach for LLMs demonstrates excellent performance, especially on the Chinese dataset. Our method further achieves the best results.


We also investigate the performance of the GPT series on three datasets, reporting the experimental results of GPT-3.5 and the latest GPT-4o versions in zero-shot and few-shot settings. Due to cost constraints, we select 200 data points from the test set for the experiments. The results indicate that on two English datasets, GPT-4o outperforms GPT-3.5 by 5.21\% and 4.31\% in F1 scores, respectively, under the 3 in-context examples setup, demonstrating stronger performance. Interestingly, however, both models perform poorly on the Chinese dataset, with GPT-4o even underperforming GPT-3.5. Analysis shows that GPT-4o extracts too much non-financial information, like person entities, indicating that its general extraction capabilities interfere with task-specific understanding.

\subsection{Ablation Study}
At this stage, we explore the contributions of various components within our framework. Table~\ref{table Ablation Study} presents the results of different model variants.

To evaluate the effectiveness of the second stage, we compare the experimental results with those from the first stage, which involves only basic fine-tuning. The results demonstrate that the correction strategy improves performance across all three datasets. 
Furthermore, we examine the role of the GNN-based example retriever in the second stage. We replace the in-context examples with the same number of fixed examples and report the corresponding results. 
Overall, the absence of any feature typically results in a decrease in F1 scores compared to the complete model. 
\begin{table}[h]
  \centering

  \begin{tabular}{lccc}
    \hline
    \multirow{2}{1cm}{Method} &\multirow{2}{1.4cm}{FinEntity} &\multirow{2}{1.4cm}{SEntFiN-Span} &\multirow{2}{1.4cm}{FinEntCN}\\\\
    \hline
    all             &89.10    &78.96    &86.75  \\
    w/o gnn         &\textcolor{DarkGreen}{(-0.84)}     &\textcolor{DarkGreen}{(-0.12)}    &\textcolor{DarkGreen}{(-0.71)} \\
    w/o stage2      &\textcolor{DarkGreen}{(-2.29)}    &\textcolor{DarkGreen}{(-0.85)}    &\textcolor{DarkGreen}{(-1.15)}  \\
    \hline
  \end{tabular}
  \caption{The Macro-F1 score(\%) of the ablation experiments.Values in green indicate the
drop in performance after removing a module.}
  \label{table Ablation Study}
\end{table}


\subsection{Key Parameters Analysis}
\label{Key Experimental Parameters Analysis}
\subsubsection{The Number of In-context Examples}

\begin{figure}[h]
  \includegraphics[width=\columnwidth]{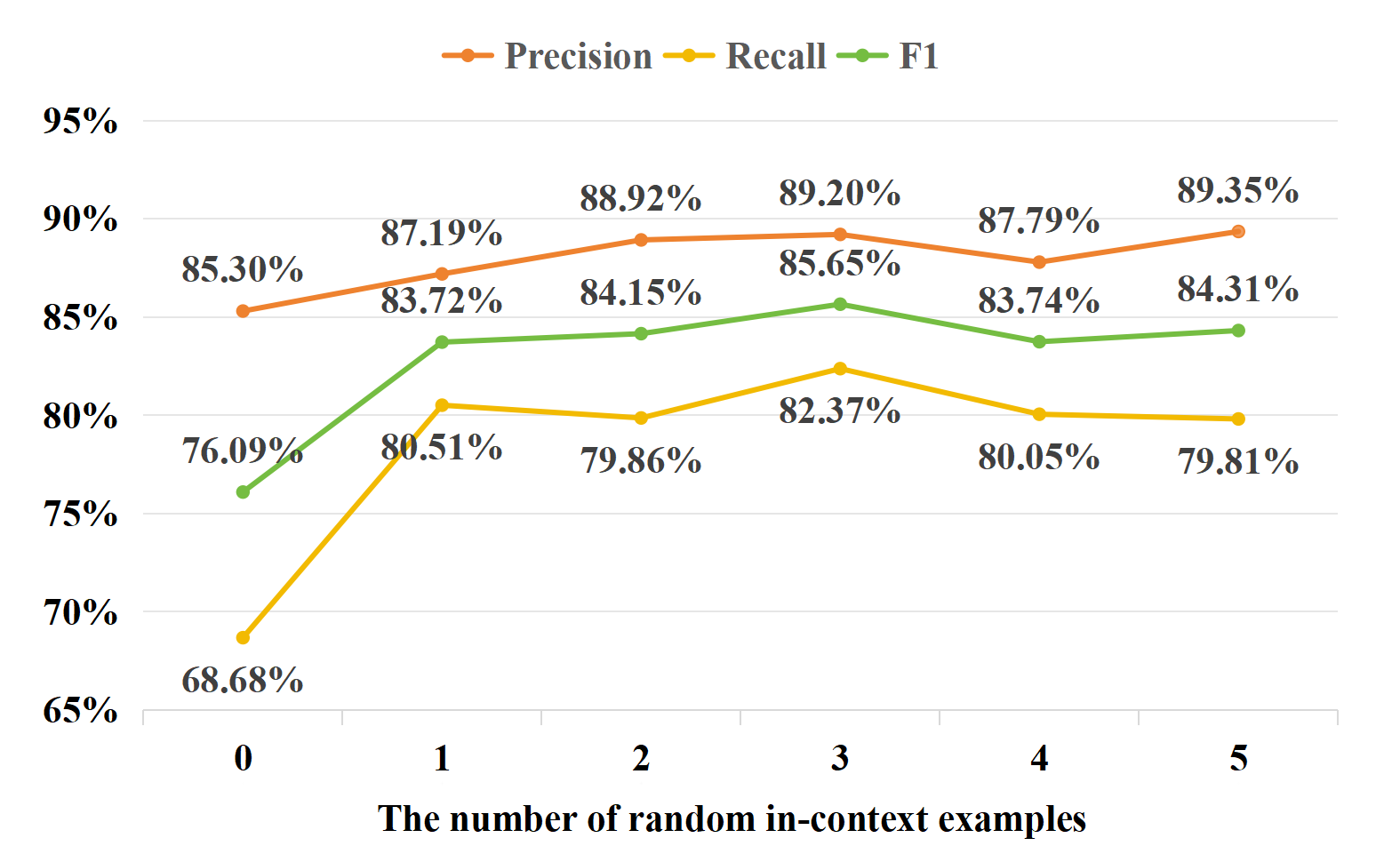}
  \caption{Performance impact of different numbers of in-context examples on the FinEntity dataset. }
  \label{table LlaMAICL-FinEntity-result}
\end{figure}

Figure~\ref{table LlaMAICL-FinEntity-result} presents the experimental results of fine-tuning the LlaMA2-7B model on the FinEntity dataset using different numbers of randomly selected in-context examples. Compared to the task fine-tuning template without examples, the inclusion of such examples leads to a significant improvement in model performance. 
As the number of in-context examples increases, the results indicate that the model achieves optimal performance with three examples. This finding provides a basis for selecting three examples in the above main experiments.

\subsubsection{The Positive-to-negative Sample Ratio}
\begin{figure}[h]
  \includegraphics[width=\columnwidth]{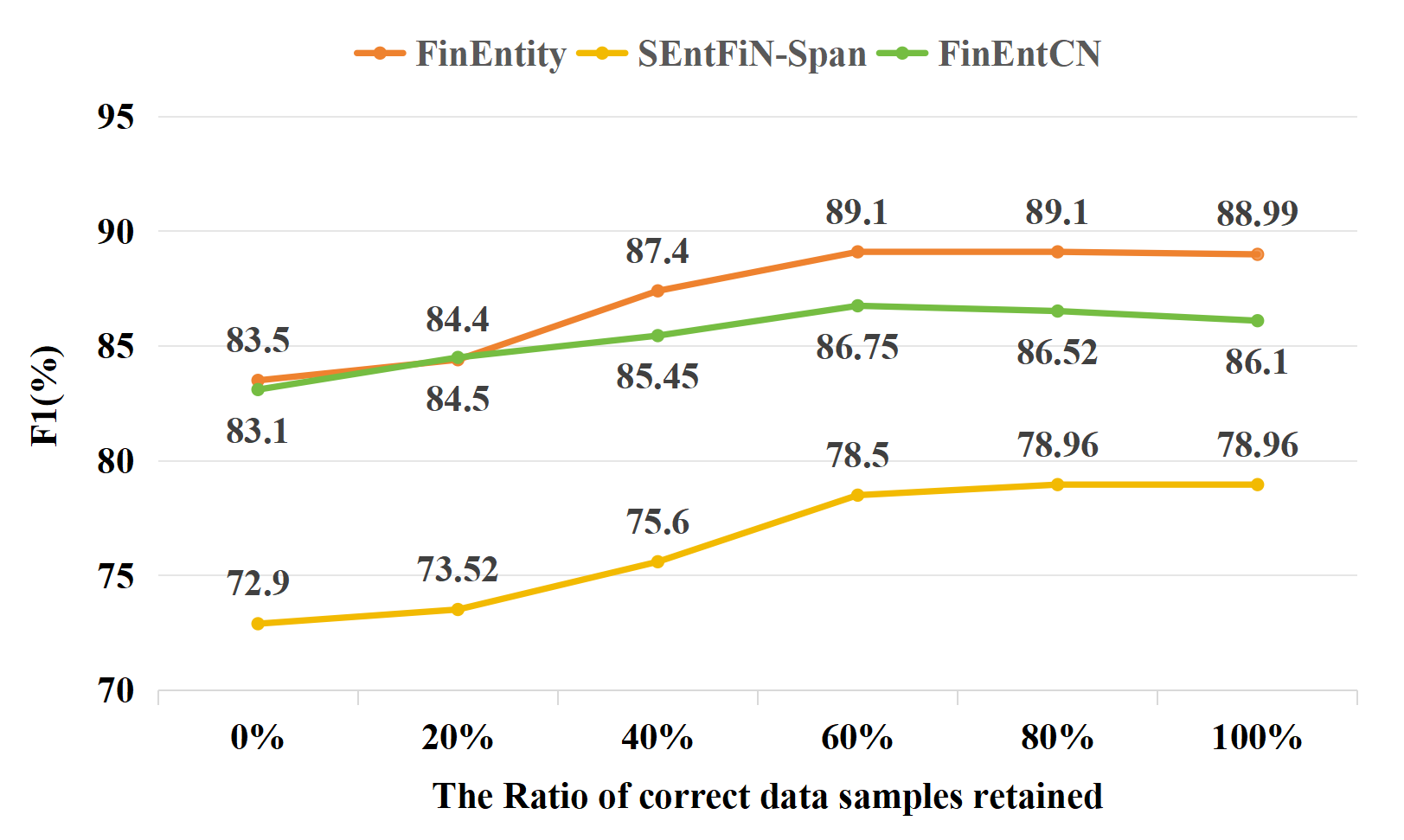}
  \caption{Impact of the ratio of correct data samples retained. }
  \label{fig:zxt}
\end{figure}

After obtaining the pseudo-labeled data from the first stage, we filter the training set and retain different proportions of correct examples to investigate the impact of the positive-to-negative sample ratio on the performance of the correction model in the second stage. The experimental results are shown in Figure~\ref{fig:zxt}. As the proportion of correct samples retained increases, the performance of the fine-tuned correction model initially improves and then declines. This indicates that when training a correction model, an excessively high or low ratio of correct to incorrect samples in the training set can negatively affect correction performance, potentially leading to worse outcomes. 

We observe that retaining about 60\% or 80\% of the correct samples yields the best performance for the correction model. At this point, the overall ratio of correct pseudo-labeled samples to incorrect pseudo-labeled samples in the training data may better align with the true accuracy, enabling the model to have a more accurate perception of its own error rate. This results in improved performance and aligns with human intuition.






\subsection{Case Study}
\label{case study}

To better demonstrate the effectiveness of our framework, we conduct case studies for both English and Chinese texts, as shown in Table~\ref{table case study} in Appendix.

In the English case 1, the model fine-tuned in the first stage accurately identifies two financial entities in the text: Twitter Inc and Tesla Inc. 
The description of Twitter Inc's stock increase is correctly classified as positive sentiment. 
However, the model incorrectly classifies the sentiment for Tesla Inc as positive as well, which is a common issue in multi-entity sentiment analysis where the model erroneously assumes uniform sentiment across multiple entities.
In the second stage, we introduce three correction examples to guide the model in evaluating and adjusting the pseudo-labels, which leads to successful results. Table~\ref{table case study2} in Appendix presents the experimental results of our method, showing the F1 scores for entity and sentiment classification. Both scores improved after our correction stage, with a greater increase in the sentiment polarity score.

Additionally, we observe that LLMs also produce significant errors similar to those in the Chinese case 1, likely due to the lack of financial domain optimization in the Chinese base models. Our correction strategy proves effective in these instances as well. English case 2 and Chinese case 2 demonstrate the limited ability of our method to correct the model's persistent misjudgments regarding entity boundaries and categories. 





\section{Case Application: Cryptocurrency Market Prediction}

Studies have shown a positive contemporaneous correlation between Bitcoin prices and entity-level sentiment scores, with the maximum information coefficient (MIC) between cryptocurrency prices and sentiment indicating a moderate positive correlation. Furthermore, entity-level sentiment demonstrates higher correlations than sequence-level sentiment~\cite{tang-etal-2023-finentity}, suggesting that market sentiment plays a positive role in regulating price volatility.

\begin{table}[h]
  \centering
  \begin{tabular}{lc}
    \hline
    Features&RMSE  \\
    \hline
    OHLC + ELS (\textbf{SILC})  &  0.07936   \\
    OHLC + ELS (Bert\_CRF)*  &  0.08502   \\
    OHLC + SLS*  & 0.09549     \\
    OHLC only*  &   0.11218  \\
    
    \hline
  \end{tabular}
  \caption{\label{table Bitcoin}
    Bitcoin price prediction performance. ELS refers to entity-level sentiment, and SLS refers to sequence-level sentiment. '*' indicates that the data is sourced from the research of ~\citet{tang-etal-2023-finentity}.
  }
\end{table}

Based on a dataset of 15,290 timestamped articles from May 20, 2022, to February 1, 2023, we conduct a Bitcoin price prediction task. Sentiments are labeled as positive (+1), neutral (0), and negative (-1), and daily sentiment scores are calculated and normalized. For the price feature, we use the Open-High-Low-Close (OHLC) price, which provides information on the market price movements during a specific time period. A long short-term memory (LSTM) network is used for prediction, with a time step of 10 and a hidden size of 128.

Table~\ref{table Bitcoin} reports the RMSE (Root Mean Squared Error) of the model predictions. RMSE measures the average error between the predicted values and the true values, with a smaller RMSE indicating higher predictive accuracy. The results indicate that the LSTM model using only OHLC prices performs the worst, while the model incorporating entity-level sentiment features outperforms both the sequence-level model and the model without sentiment features. The sentiment score features derived from SILC method achieve the best performance.










\section{Conclusions}
Our research focuses on the task of entity-level sentiment analysis in the financial domain, for which we have constructed the largest English and Chinese datasets. 
Moreover, we propose an innovative strategy called "Self-aware In-context Learning Correction" (SILC). The SILC framework consists of two stages and significantly improves the accuracy by enabling the model to learn correction examples relevant to the current instance. Experimental results demonstrate that the proposed SILC strategy effectively enhances model performance, achieving state-of-the-art results. Additionally, the case study in the cryptocurrency market demonstrates the practical utility of our datasets and methods, which we believe are valuable resources for financial sentiment analysis.





\section*{Limitations}
The proposed method involves multiple training stages, which, while enhancing model refinement, also increase training time and computational requirements. This could impact scalability and resource use. To mitigate these issues, optimization techniques such as model pruning and knowledge distillation, as well as cloud computing and distributed training, can be employed.


\section*{Ethics Statement}

This study adheres to the~\href {https://www.aclweb.org/portal/content/acl-code-ethics}{ACL Code of Ethics}. The data collected comes entirely from publicly accessible datasets. Our constructed datasets do not disseminate personal information and do not contain content that could potentially harm any individual or community. 


\section*{Acknowledgments}

The authors thank the anonymous reviewers for their insightful comments. This work is mainly supported by the Key Program of the Natural Science Foundation of China (NSFC) (No.U23A20316), the Key R\&D Project of Hubei Province (No.2021BAA029), the Key Research and Development Program of Henan Province (No.241111212700) and the Key Lab of Information Network Security, Ministry of Public Security (No.C23600-04).


\bibliography{custom}

\appendix

\section{Instruction Template}
\label{Appendix A}
\begin{CJK}{UTF8}{gbsn}

\subsection{The Instruction Template for ChatGPT}
This section provides a directive template for utilizing the GPT model API, designed to perform entity-level financial sentiment analysis and consisting of four key components: System Message, Guidelines, Examples, and Task Text.


\begin{figure}[h]
  \includegraphics[width=\columnwidth]{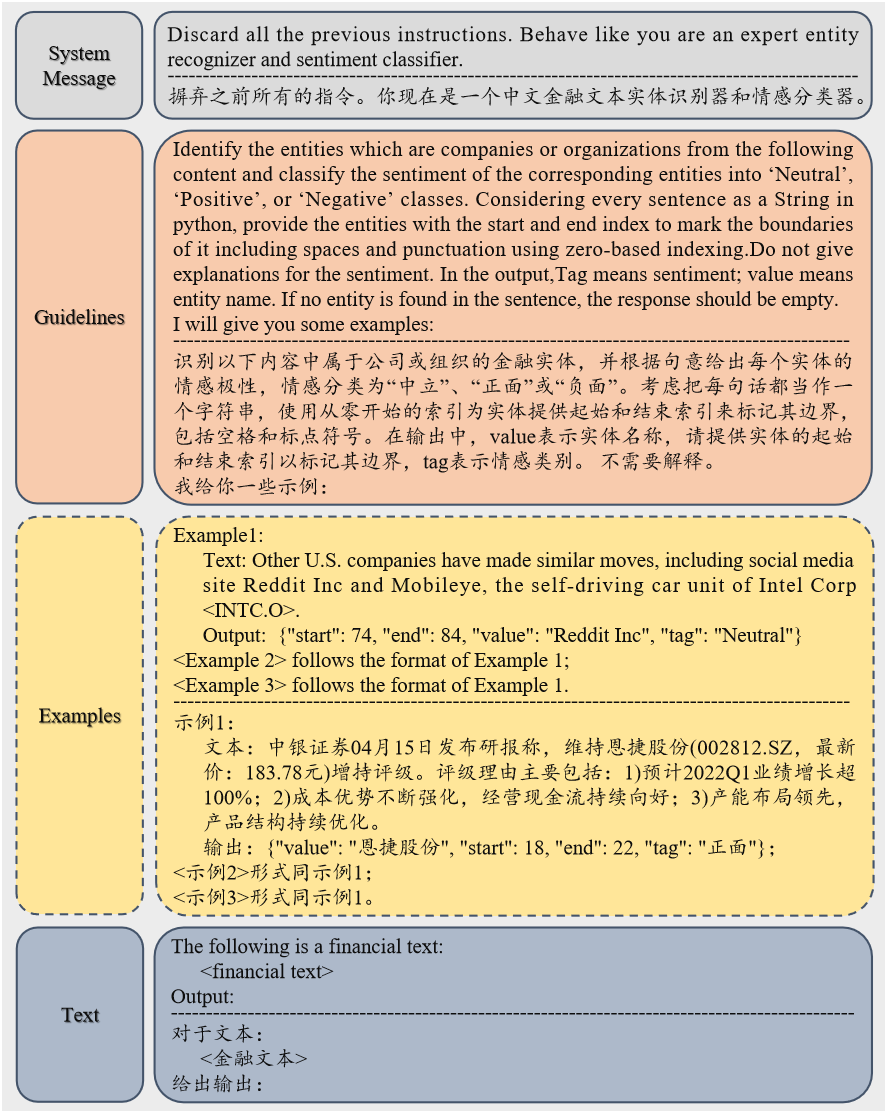}
  \caption{The English and Chinese instruction template for ChatGPT. }
  \label{fig:GPTPrompt}
\end{figure}

\subsection{The Instruction Template for SILC}
This section introduces the fine-tuning instruction templates for the two stages of SILC. We design multiple similar fine-tuning instruction templates, and one is randomly selected during use. One of them is presented here as an example.

\begin{figure}[h]
  \includegraphics[width=\columnwidth]{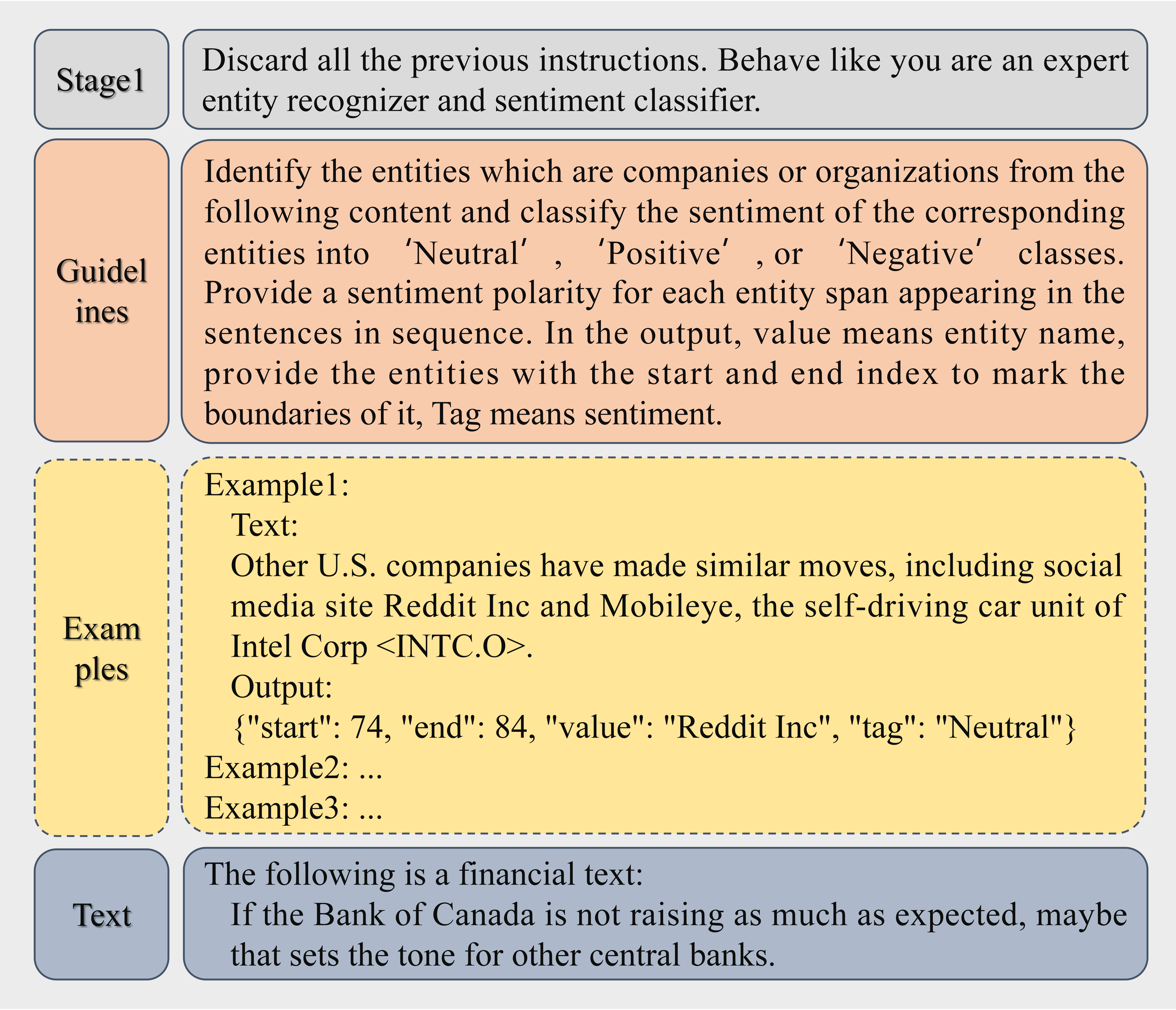}
  \caption{Fine-tuning instruction template for the initial response generation. }
  
  \label{fig:stage1prompt}
\end{figure}





\begin{figure}[h]
  \includegraphics[width=\columnwidth]{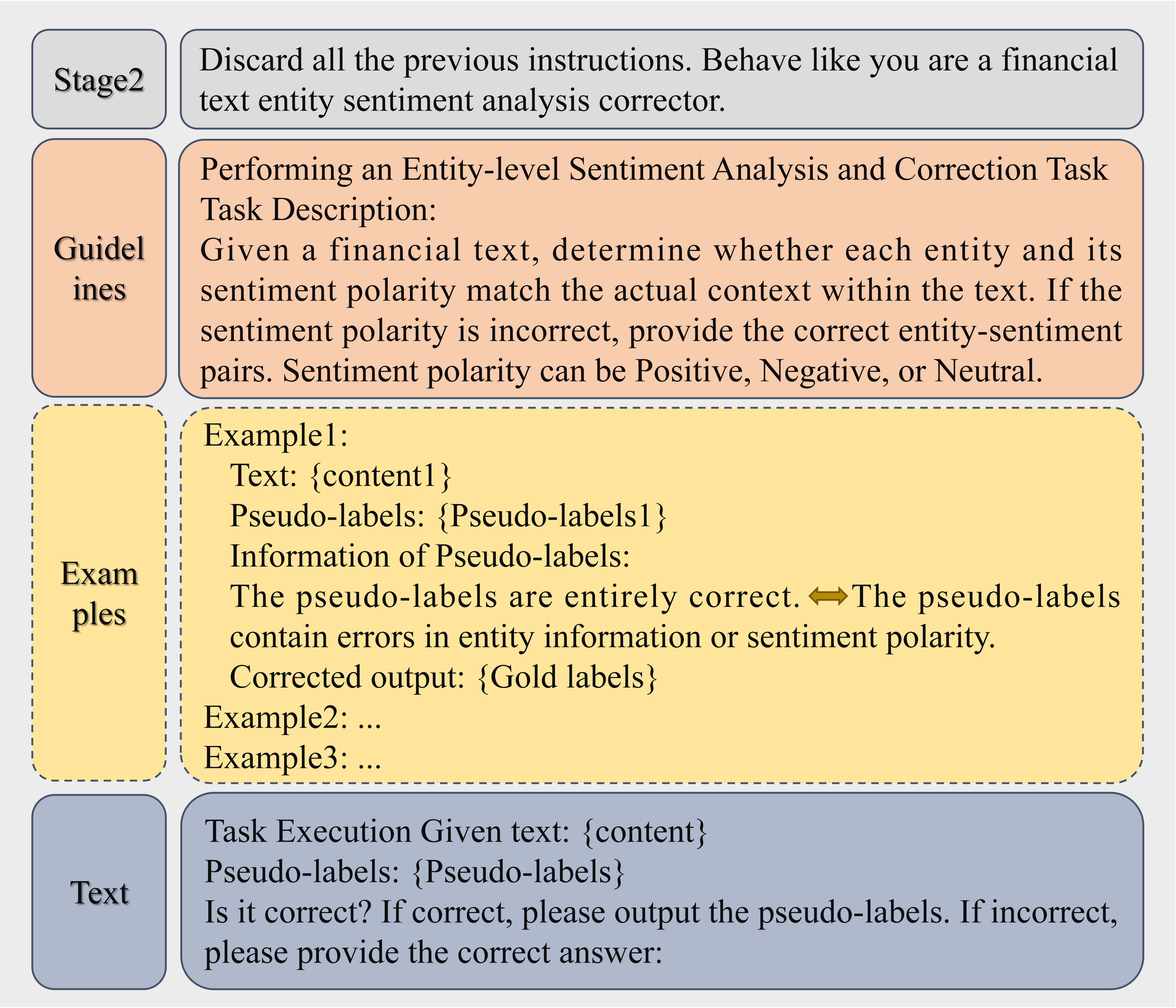}
  \caption{Fine-tuning instruction template for the self-correction steps. The bidirectional arrow indicates that the real usage is either the preceding or the following.}
  
  \label{fig:stage2prompt}
\end{figure}











\end{CJK}

\section{Experimental Parameter Settings}
\label{Appendix B}

\textbf{Our method} involves multiple stages. In the first stage of supervised fine-tuning, the learning rate and number of epochs are set to $8 \times 10^{-5}$ and 5, respectively. For the GNN training phase, we define different heuristic rules for $\theta_{Lig}$ and $\theta_{Sen}$ to distinguish between linguistic and sentiment feature similarities. The linguistic ($\theta_{Lig}$) and sentiment ($\theta_{Sen}$) parameters are illustrated in Table~\ref{table gnn set}. We use a BERT-based uncased tokenizer as the token encoder, with an initial learning rate of $1 \times 10^{-4}$, running for 10 epochs. In the fine-tuning stage of the correction model, we insert the top 3 most relevant examples (k = 3), ranked by similarity~\cite{liu-etal-2022-makes}, including one linguistic example, one sentiment example, and one average sample. We leverage Low-Rank Adaptation (LoRA)\cite{hulora} for efficient parameter tuning. All methods utilize the AdamW optimizer\cite{loshchilov2019decoupledweightdecayregularization}, incorporating gradient decay, dynamic learning rate scheduling, and gradient clipping techniques.

\begin{table}[h]
  \centering
  \begin{tabular}{l c|c|c}
    \hline
    &FinEntity 	&SEntFiN-Span	&FinEntCN  \\
    \hline
    {$\theta_{Lig}$}  &0.37  &0.46   &0.15    \\
    {$\theta_{Sen}$}  &0.8   &0.8   &0.7      \\
    \hline
  \end{tabular}
  \caption{\label{table gnn set}
    The gnn retrieval model training parameters.
  }
\end{table}
All experiments are conducted on an Ubuntu 20.04.5 server equipped with a V100-32G GPU. We randomly split 10\% of the training set as the validation set and select the best-performing model, using Macro-F1 as the primary evaluation metric. Each experiment is repeated three times with different random seeds, and we report the average results.

\textbf{BGCA Method Experimental Settings: }

BGCA is a data augmentation strategy based on the T5 model. Originally a cross-domain data augmentation method, it is modified to enhance the training set specifically within the financial domain.
The experimental parameters are set as follows: 
num\_train\_epochs is set to 15, learning\_rate is $3 \times 10^{-4}$, data\_gene\_epochs is 20, and whether to use the same model is set to use\_same\_model.

\textbf{SpanABSA Method Experimental Settings: }
The task is set to run\_joint\_span, train\_batch\_size is 8, and logit\_threshold is 8.0.

\textbf{InstructABSA Method Experimental Settings:}
num\_train\_epochs is set to 4 and learning\_rate is $5 \times 10^{-5}$.

Task prompt template: """Definition: The output will include each financial entity (both implicit and explicit) along with their sentiment polarity. If there are no financial entities, the output should be noaspectterm: none. The output should follow the order in which the financial entities appear in the text.
Two positive examples, 
Two negative examples, 
Two neutral examples, 
Now complete the following example
input: """





\section{Case Study}

\begin{CJK}{UTF8}{gbsn}

\begin{table*}[hbpt]
  \centering

  \begin{tabular}{|p{1.2cm}|p{2.8cm}|p{11cm}|}
    \hline
\multicolumn{3}{|c|}{English case study} \\ \hline

\multirow{4}{*}{Case\#1}
     &Sentence &Micro-blogging site \textcolor{DarkBlue}{Twitter Inc} <TWTR.N> gained 3.8\%, adding to its 27\% surge in the previous session, after saying it will name top shareholder and \textcolor{DarkBrown}{Tesla Inc} <TSLA.O> CEO Elon Musk to its board. \\  \cdashline{2-3}
     
     &Gold labels &\{value: \textcolor{DarkBlue}{Twitter Inc}, start: 20, end: 31, tag: \textcolor{DarkBlue}{Positive}\}\newline\{value: \textcolor{DarkBrown}{Tesla Inc}, start: 149, end: 158, tag: \textcolor{DarkBrown}{Neutral}\} \\ \cdashline{2-3}
     
    &Stage1 Predict\newline(Pseudo-labels) &\{value: \textcolor{DarkBlue}{Twitter Inc}, start: 20, end: 31, tag: \textcolor{DarkBlue}{Positive}\} \textcolor{DarkGreen}{\ding{51}} \newline\{value: \textcolor{DarkBrown}{Tesla Inc}, start: 149, end: 158, tag: \textcolor{DarkBrown}{Positive}\}\ding{55}   \\ \cdashline{2-3}
    &Stage2 Predict\newline(Corrected-labels) &\{value: \textcolor{DarkBlue}{Twitter Inc}, start: 20, end: 31, tag: \textcolor{DarkBlue}{Positive}\}\textcolor{DarkGreen}{\ding{51}}\newline\{value: \textcolor{DarkBrown}{Tesla Inc}, start: 149, end: 158, tag: \textcolor{DarkBrown}{Neutral}\}\textcolor{DarkGreen}{\ding{51}} \\  \hline

    \multirow{4}{*}{Case\#2}
    
     &Sentence &\textcolor{DarkYellow}{Rupiah} leads Asia FX losses after solid US data, weekly slides seen. \\ \cdashline{2-3}

      &Gold labels&\{value: \textcolor{DarkYellow}{Rupiah}, start: 0, end: 6, tag: \textcolor{DarkYellow}{negative}\} \\ \cdashline{2-3}
      
      &Stage1 Predict\newline(Pseudo-labels) &\{value: \textcolor{DarkYellow}{Rupiah}, start: 0, end: 6, tag: \textcolor{DarkYellow}{negative}\}\textcolor{DarkGreen}{\ding{51}}\newline \{value: Asia FX, start: 13, end: 20, tag: negative\}\ding{55}\\ \cdashline{2-3}
        &Stage2 Predict\newline(Corrected-labels) &\{value: \textcolor{DarkYellow}{Rupiah}, start: 0, end: 6, tag: \textcolor{DarkYellow}{negative}\}\textcolor{DarkGreen}{\ding{51}}\newline \{value: Asia FX, start: 13, end: 20, tag: negative\}\ding{55} \\  \hline
        
\multicolumn{3}{|c|}{Chinese case study} \\ \hline
    \multirow{4}{*}{Case\#1}
    
     &Sentence & 2月3日,\textcolor{DarkBlue}{荣联科技}(维权)收深交所关注函。公司此前披露,因涉嫌信息披露违法违规,证监会决定对公司立案。深交所要求公司说明非公开发行股票进展情况,并说明立案调查对公司非公开发行股票事项的影响。值得注意的是,2月3日当日,\textcolor{DarkBrown}{荣联科技}还录得涨停。 \\ \cdashline{2-3}

&Gold labels &\{value: \textcolor{DarkBlue}{荣联科技}, start: 5, end: 9, tag: \textcolor{DarkBlue}{负面}\}\newline\{value: \textcolor{DarkBrown}{荣联科技}, start: 120, end: 124, tag: \textcolor{DarkBrown}{正面}\}\\ \cdashline{2-3}
     
      &Stage1 Predict\newline(Pseudo-labels) &\{value: \textcolor{DarkBlue}{荣联科技}, start: 5, end: 9, tag: \textcolor{DarkBlue}{中立}\}\ding{55}\newline\{value: \textcolor{DarkBrown}{荣联科技}, start: 120, end: 124, tag: \textcolor{DarkBrown}{中立}\}\ding{55}  \\ \cdashline{2-3}
        &Stage2 Predict\newline(Corrected-labels) &\{value: \textcolor{DarkBlue}{荣联科技}, start: 5, end: 9, tag: \textcolor{DarkBlue}{负面}\}\textcolor{DarkGreen}{\ding{51}}\newline\{value: \textcolor{DarkBrown}{荣联科技}, start: 120, end: 124, tag: \textcolor{DarkBrown}{正面}\}\textcolor{DarkGreen}{\ding{51}}\\  \hline

    \multirow{4}{*}{Case\#2}
    
     &Sentence & \textcolor{DarkYellow}{东风集团}股份在港交所公告，2021年12月销售乘用车227782辆，上年同期为293747辆；全年累计销售乘用车2252496辆，同比下降2.6\%。其中，2021年12月销售新能源汽车26383辆，上年同期为12661辆；全年累计销售新能源汽车160641辆，同比增长263.3\%。
 \\ \cdashline{2-3}

&Gold labels &\{value: \textcolor{DarkYellow}{东风集团}, start: 0, end: 4, tag: \textcolor{DarkYellow}{负面}\} \\  \cdashline{2-3}

      &Stage1 Predict\newline(Pseudo-labels) &\{value: \textcolor{DarkYellow}{东风集团股份}, start: 0, end: 6, tag: \textcolor{DarkYellow}{负面}\}\ding{55} \\ \cdashline{2-3}
        &Stage2 Predict\newline(Corrected-labels) &\{value: \textcolor{DarkYellow}{东风集团股份}, start: 0, end: 4, tag: \textcolor{DarkYellow}{负面}\}\ding{55} \\  \hline

  \end{tabular}
  \caption{English and Chinese case studies.}
  \label{table case study}
\end{table*}

\begin{table*}[hbpt]
  \centering 
  \begin{tabular}{l c|c|c|c|c|c}
    \hline
    \multirow{2}{*}{Method}    & \multicolumn{2}{c|}{FinEntity} & \multicolumn{2}{c|}{SEntFiN-Span} & \multicolumn{2}{c}{FinEntCN}\\ \cline{2-7}
    
    &Entity	&Sentiment	&Entity	&Sentiment &Entity	&Sentiment\\\hline
stage1&0.9415  	&0.9469	&0.8960	&0.8664   &0.9321  	&0.9201\\  \hline
stage2&0.9433  	&0.9563	&0.8994	&0.8758   &0.9382  	&0.9308\\  \hline

  \end{tabular}
  \caption{The F1 scores for entity and sentiment polarity in two stages on our dataset. The calculation method deems a prediction correct as long as the predicted entity or sentiment appears in the ground truth, without requiring sequence or full matching.}

  \label{table case study2}
\end{table*}

\end{CJK}

\end{document}